\documentclass[10pt,twocolumn,letterpaper]{article}
\usepackage{iccv}              % To produce the CAMERA-READY version
\definecolor{iccvblue}{rgb}{0.21,0.49,0.74}
\usepackage[pagebackref,breaklinks,colorlinks,allcolors=iccvblue]{hyperref}

\usepackage{algorithm}
\usepackage{algorithmic}
\usepackage{multirow}
\usepackage{pifont}
\usepackage{xcolor}
\def\paperID{10713} % *** Enter the Paper ID here
\def\confName{ICCV}
\def\confYear{2025}

\title{DiffuseSlide: Training-Free High Frame Rate Video Generation Diffusion}

\author{
Geunmin Hwang$^{1, 3}$ \quad
Hyun-kyu Ko$^{4}$ \quad
Younghyun Kim$^{2}$ \quad
Seungryong Lee$^{4}$ \quad
Eunbyung Park$^{2}$\textsuperscript{\dag}\\
\\
$^{1}$RECON Labs Inc. \\
$^{2}$Department of Artificial Intelligence, Yonsei University \\
$^{3}$Department of Artificial Intelligence, Sungkyunkwan University \\
$^{4}$Department of Electrical and Computer Engineering, Sungkyunkwan University\\
\href{https://geunminhwang.github.io/DiffuseSlide/}{\textnormal{geunminhwang.github.io/DiffuseSlide/}}
}
\begin{document}

\maketitle

\begin{abstract}
Recent advancements in diffusion models have revolutionized video generation, enabling the creation of high-quality, temporally consistent videos. However, generating high frame-rate (FPS) videos remains a significant challenge due to issues such as flickering and degradation in long sequences, particularly in fast-motion scenarios. Existing methods often suffer from computational inefficiencies and limitations in maintaining video quality over extended frames. In this paper, we present a novel, training-free approach for high FPS video generation using pre-trained diffusion models. Our method, \textit{DiffuseSlide}, introduces a new pipeline that leverages key frames from low FPS videos and applies innovative techniques, including noise re-injection and sliding window latent denoising, to achieve smooth, consistent video outputs without the need for additional fine-tuning. Through extensive experiments, we demonstrate that our approach significantly improves video quality, offering enhanced temporal coherence and spatial fidelity. The proposed method is not only computationally efficient but also adaptable to various video generation tasks, making it ideal for applications such as virtual reality, video games, and high-quality content creation.

\end{abstract}    
\section{Introduction}
\label{sec:intro}

Recent advancements have markedly enhanced the capabilities of video generation technology, producing high-quality, temporally coherent sequences. By leveraging large-scale datasets, advanced neural network architectures, and sophisticated training techniques, current state-of-the-art diffusion video models can generate realistic, contextually rich video sequences that exhibit both spatial and temporal fidelity~\cite{openai2024sora}. This unprecedented progress establishes them as valuable tools for various applications, such as film production and immersive AR/VR environments.

While successful and promising, current video generation models often struggle with quality degradation as the length of the video increases, with visual artifacts and inconsistencies becoming more pronounced over time. This challenge is especially critical in high-tempo scenarios, which demand seamless transitions and smooth visual experiences to maintain realism and viewer immersion. Consequently, there is an increased need to improve the current technology for high frame-rate video generation, particularly for fast-moving scenes where lower frame rates lead to choppy and unsatisfactory visual quality.

One intuitive approach for generating high frame-rate video is keyframe interpolation. By creating intermediate frames, these approaches demonstrate smoother transitions between keyframes. Traditional methods, such as optical flow~\cite{bao2019depth,niklaus2020softmax, jeong2024ocai} and kernel-based approaches~\cite{lu2022video, zhang2023extracting}, estimate motion between frames to synthesize intermediate frames. Since optical flow computes pixel-level motion vectors to guide frame synthesis, it sometimes falls short in maintaining visual quality, particularly in videos with complex or large-scale motion patterns. Similarly, kernel-based methods, which rely on local convolutions, encounter limitations when faced with substantial spatial displacements.

Recently, numerous studies~\cite{jain2024video, lyu2024frame} have explored the use of generative models for frame interpolation, with diffusion models emerging as particularly promising. For instance, LDMVFI~\cite{danier2024ldmvfi} and MCVD~\cite{voleti2022mcvd} treat frame interpolation as a conditional generation task, using the diffusion process to synthesize intermediate frames between keyframes. These models provide higher visual fidelity and temporal consistency, particularly in scenarios involving complex motion, outperforming traditional pixel-based methods in terms of quality and robustness.

Despite these advances, most existing video generation and frame interpolation models face challenges when scaling to high frame rates due to computational and memory constraints.
In this work, we propose a novel method for high frame-rate video generation using a pre-trained video generation diffusion model without any additional training or fine-tuning. We introduce a new high frame-rate video generation pipeline, \textit{DiffuseSlide}, which leverages low frame-rate keyframes as a condition to generate interpolated frames. Inspired by recent studies, our proposed pipeline incorporates noise-denoising, noise re-injection, and a sliding window approach for multi-image conditioning. First, we generate low frame-rate video latents using a pre-trained image-to-video diffusion model. These latents are then linearly interpolated in latent space to generate initial high frame-rate latents. Second, to mitigate artifacts in the interpolated frames, we introduce controlled noise into the initial latent, partially disrupting its structure to enable refinement. Finally, a reverse diffusion process with noise re-injection is applied to denoise the frames, resulting in smooth and high-quality videos with enhanced temporal consistency.

An alternative training-free approach~\cite{yang2024zerosmooth} reinterprets frame interpolation as a video restoration task, employing DDNM’s~\cite{wang2022zero} null space projection method to align interpolated frames with the space of the keyframes. Although this approach shows promise, this method has a distinct limitation: if the quality of the generated keyframes is poor, these imperfections are directly transferred to the interpolated frames, compromising overall quality. In contrast, our approach refines both the interpolated frames and the generated keyframes during the denoising stage, thereby ensuring that all frames are improved together, resulting in a higher-quality and more temporally consistent video.

\begin{figure*}[ht!]
    \centering
    \includegraphics[width=0.95\textwidth]{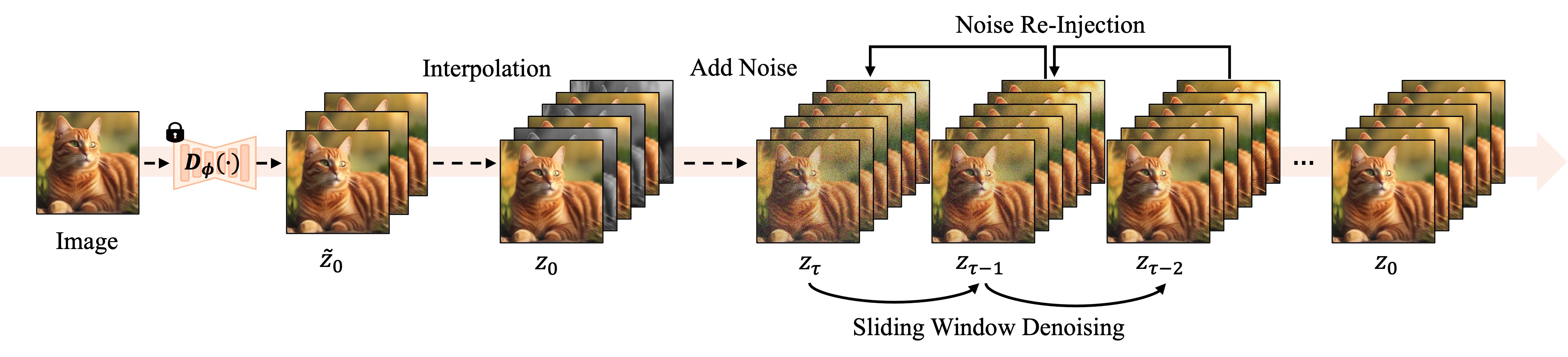} 
    \caption{Overall pipeline of \textit{DiffuseSlide}. Given an input image, a low frame-rate video is generated using a pre-trained image-to-video model. To increase the frame rate, linear interpolation is applied, followed by noise injection and denoising to remove artifacts. Our method employs multi-step noise reinjection to better align with the data manifold. Our sliding window-denoise approach conditions on multiple frames to better align with the initial low frame-rate video.}
    \label{fig:fig_pipeline}
    \vspace{-5.0pt}
\end{figure*}

Leveraging a pre-trained denoising U-Net offers the advantage of generating high-quality frames without requiring additional training.
However, the limited capacity of the pre-trained denoising U-Net can lead to blurring and over-saturation in later frames, particularly when processing a large number of interpolated frames. To handle long video sequences, previous works~\cite{qiu2023freenoise, yang2024zerosmooth} suggested rescheduling the whole denoising process and applying attention consecutively within a length manageable by diffusion models. However, this approach restricts conditioning to only the first keyframe, resulting in insufficient conditioning for the latter frames. To address this limitation, we introduce a sliding window approach that divides the entire latent sequence into manageable subsequences, each conditioned on its respective keyframes and denoised independently. Fig.~\ref{fig:fig_pipeline} provides an overview of the overall pipeline of our method.

We evaluated our method on the WebVid-10M~\cite{bain2021frozen}. Compared to other baseline methods, our approach achieves state-of-the-art (SOTA) performance across widely used metrics, including FVD, PSNR, and SSIM. The contributions of our work are summarized as follows:

\begin{itemize}
    \item We propose a novel training-free high frame-rate video generation pipeline, \textit{DiffuseSlide}, which leverages low frame-rate key frames as conditions to generate interpolated frames, producing high frame-rate videos without additional training or fine-tuning.
    
    \item Our method introduces a sliding window approach in the denoising process, enabling multi-image conditioning that enhances both temporal and spatial consistency across the generated frames.
    
    \item Through extensive experiments on the WebVid-10M Dataset and ablation studies, we demonstrate the superiority and adaptability of our method, achieving state-of-the-art (SOTA) results on key metrics, including FVD, PSNR, and SSIM.
\end{itemize}

\section{Related Works}
\vspace{-1mm}

\paragraph{Video diffusion models.}
Building on the success of the text-to-image diffusion models~\cite{rombach2022high, podell2023sdxl, saharia2022photorealistic, ramesh2022hierarchical, betker2023improving}, video diffusion models (VDMs), which model the spatio-temporal distribution of video, have emerged as a dominant approach in the realm of generative video generation. Jointly synthesizing the consecutive frames, VDMs have recently demonstrated remarkable capabilities in generating visually appealing and high-quality video samples. 

The first VDM~\cite{ho2022video} utilized 3D U-Net, which is factorized over the space and time, modeled the video in pixel space. Other 
works~\cite{ho2022imagen, singer2022make} adopted cascade scheme to generate higher-resolution videos at the  pixel levels. Subsequently, several studies~\cite{he2022latent,zhang2023i2vgen, gupta2025photorealistic, xing2025dynamicrafter} proposed modeling the videos in the latent space to enhance the computational efficiency during training and inference. Many other works~\cite{wang2023modelscope, hong2022cogvideo, villegas2022phenaki, blattmann2023align, chen2023videocrafter1, chen2024videocrafter2, wang2023lavie, blattmann2023stable, kong2024hunyuanvideo}, either conditioned on text or image prompts, followed this latent-based paradigm and further advance the field. 

Despite previous advancements, computational challenges inherent to video data have constrained existing methods to generating short clips, typically of 16 or 24 frames. This limitation in video diffusion model capacity often leads to unnatrual transitions between frames and choppy outputs, impeding various practical applications.

\vspace{-4mm}
\paragraph{Video frame interpolation.}
A naive solution to the aforementioned low frame-rate issue is to train or fine-tune the base video diffusion models using videos with larger number of frames~\cite{yang2024cogvideox,openai2024sora}. However, such approach is formidable for most research institutions, since it requires substantial computational resources, therefore primarily limited to a selected few organizations. As a result, the adaptation of video diffusion models for producing high frame-rate videos has not been sufficiently explored. Recently, ZeroSmooth~\cite{yang2024zerosmooth} proposed a training-free pipeline capable of producing higher frame rate videos. By incorporating hidden state correction modules, which work in a plug-and-play manner within the transformer block of the video diffusion model, their method enables the base video diffusion model to produce visually smooth videos. 

In this paper, we propose a training-free high frame-rate video generation pipeline, dubbed \textit{DiffuseSlide}, which enables video diffusion models to generate seamless, smooth, and temporally coherent videos. Our proposed \textit{DiffuseSlide} operates without modifying the architecture or weights of the original model, thereby not only fully leveraging the powerful generative abilities of the base model but also making it easily adaptable to any video diffusion models.
\section{Method}
\vspace{-1mm}

\subsection{Preliminary}
\vspace{-1mm}
%------------------------[Please check if there's any error...]------------
We provide a concise summary of latent video diffusion model in this section. We utilize the variance exploding (VE)~\cite{song2020score} diffusion formulation throughout the paper.

Let $p_{data}(x)$ be the video data distribution and $x_0 \in \mathbb{R}^{3 \times f \times H \times W}$ the video sample sampled from $p_{data}(x)$, i.e., $x_0 \sim p_{data}(x)$. Then, the latent video diffusion model first encodes $x_0$ into the compact latent space variable $z_0 \in \mathbb{R}^{c \times f \times h \times w}$ for computational efficiency. Let $p(z; \sigma_t)$ denote the marginal probability of noisy latent $z_t = z_0 + \sigma_t\epsilon$, where $\epsilon \sim \mathcal{N}(0, I)$. Note that for sufficiently large $\sigma_T = \sigma_{max}$, the noisy latent $z_T$ is indistinguishable from the pure Gaussian noise of variance $\sigma_{max}^2$, i.e., $p(z; \sigma_{T}) \approx \mathcal{N}(0, \sigma_{max}^2)$.
%-----------------------------------------------------------------------------

Starting from the high-variance Gaussian noise $z_T \sim \mathcal{N}(0, \sigma_{max}^2)$, the diffusion model $D_{\phi}(\cdot)$ parametrized by $\phi$ learns to gradually denoise it towards the clean video latent distribution. Finally, the obtained clean latent $z_0$ is decoded back to the clean video sample via the latent decoder.
\subsection{Problem Formulation}
\vspace{-1mm}
%-------------------------[After]-------------------------------------------------------
In this study, we propose a novel, training-free approach for high frame-rate video generation by levarging image-to-video diffusion models~\cite{blattmann2023stable, zhang2023i2vgen}. Formally, given pre-trained video diffusion model $D_{\phi}(\cdot)$, trained on a set of fixed-length frames $\tilde{x}_0 \in \mathbb{R}^{3 \times f \times H \times W}$, and condition image $y$, our aim is to generate high frame-rate video $x_0 \in \mathbb{R}^{3 \times F \times H \times W}$ where $f \ll F$, without modifying or training $\phi$.
%--------------------------------------------------------------------------

\subsection{High Frame Rate-Frame Interpolation}
\vspace{-1mm}
Our high frame-rate image-to-video generation pipeline starts by generating a low frame rate latent $\tilde{z}_0 \in \mathbb{R}^{c \times f \times h \times w}$ using the pre-trained image-to-video model, where typically $h = \frac{H}{8}, w = \frac{W}{8}$. The frames generated at this stage act as `keyframes', providing a structural foundation for the subsequent steps.

To increase the frame rate, we apply linear interpolation between frames in the generated latent $\tilde{z}_0$, generating intermediate frames that serve as initialization for the latter process:
\begin{equation}
    z_0 = \texttt{INTERP}(\tilde{z}_0) \in \mathbb{R}^{c \times F \times h \times w}.
    \label{eq:eq1}
\end{equation}

Since the intermediate frames are obtained by simply averaging neighboring frames, the initialized video contains flickering artifacts and overlapping ghosting effects.

To address this, we add noise~\cite{meng2021sdedit, kim2024diffusehigh, guo2025make} corresponding to the diffusion timestep $\tau < T$ to ${z}_0$, intentionally disrupting the spatiotemporal structure of the initial video to facilitate the reconstruction of high-frequency details.
\begin{equation}
    z_\tau = z_0 + \sigma_\tau^2\epsilon, \quad \epsilon \sim \mathcal{N}(0, I),
    \label{eq:eq2}
\end{equation}
where $\sigma_\tau^2$ is the variance of the Gaussian noise at timestep $\tau$.
The noise level must be carefully balanced: too much noise can excessively distort the structure, even disrupting the keyframes and leading to a completely different reconstruction, while too little noise may leave artifacts unresolved.

\begin{figure}[h!]
    \centering
    \includegraphics[width=0.95\columnwidth]{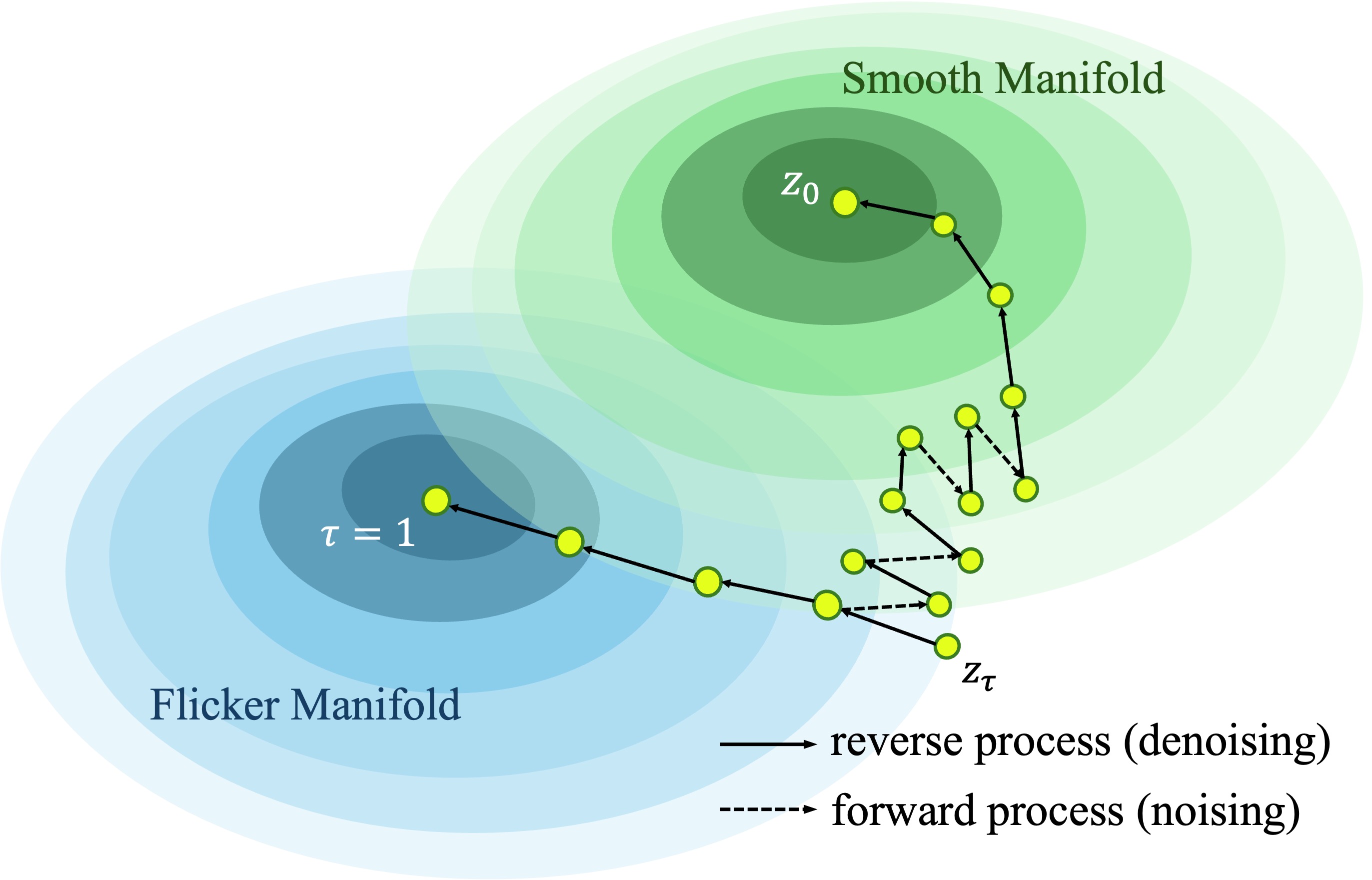} 
    \caption{Noise re-injection adds flexibility to manifold constraints, enabling jumps across manifolds. Since the starting point of our denoising process may slightly deviate from the smooth video manifold, we inject noise after each denoising step. This approach helps to break down artifacts and reconstruct details that better align with the smooth video manifold.}
    \label{fig:fig_renoise}
\end{figure}

\subsection{Noise Re-Injection for Extra Flexibility}
\vspace{-1mm}
In the previous stage, we added an appropriate level of noise to disrupt the spatiotemporal structure of the initial video, setting it up for high-frequency detail reconstruction. However, following the standard denoise scheduling can be insufficient for producing smooth and coherent video, as the level of degradation in the interpolated video often differs from what the pre-trained model is optimized to handle at each denoising step. To address this discrepancy—where a single denoising step frequently fails to align interpolated frames with keyframes under the model’s noise schedule—we incorporate a noise re-injection strategy inspired by recent studies in bounded generation techniques for diffusion models~\cite{feng2024explorative, yang2024vibidsampler, wang2024generative}.

We implement a controlled re-injection of noise, defined as follows,
\begin{equation}
    z_{\tau-\delta} = \texttt{Reinject}(z_{\tau}, \delta).
    \label{eq:alg1}
\end{equation}
Specifically, noise corresponding to timestep $\tau$ is first added to the interpolated latent and then denoised; afterward, a small amount of noise $\epsilon$ is re-injected. This alternating process of denoising and noise re-injection is performed $M$ times at each timestep, enhancing the flexibility required for matching temporal consistency between interpolated and keyframes. Please refer to (Alg.~\ref{alg:noise_reinjection}) for further details.

This iterative noise re-injection strategy allows for additional refinement, progressively guiding the interpolated frames closer to the pre-trained model’s data manifold. By repeating this process across $\delta$ timesteps, the interpolated frames gradually converge toward the data manifold, yielding a smoother, temporally coherent video. As shown in Fig.~\ref{fig:fig_renoise}, this process effectively mitigates temporal inconsistencies by preventing the model from being confined to suboptimal trajectories.

\begin{algorithm}[t]
    \caption{SlidingWindow-Denoise-Reinject}
    \label{alg:noise_reinjection}
    \textbf{Input}: Interpolated latent $z_{\tau} \in \mathbb{R}^{c \times F \times h \times w}$ at timestep $\tau$
    \textbf{Output}: Denoised latent with noise reinjection, $z_{\tau - 1}$

    \begin{algorithmic}[1]
        \FOR{$m = 1$ to $M$}
            \STATE $z_{\tau-1} \gets \text{SlidingWindow-Denoise}(z_{\tau})$ 
            \STATE $\epsilon \sim \mathcal{N}(0, \sqrt{\sigma_{\tau}^2 - \sigma_{\tau-1}^2}I)$
            \STATE $z_{\tau} \gets z_{\tau-1} + \epsilon$ 
        \ENDFOR
        \STATE $z_{\tau-1} \gets \text{SlidingWindow-Denoise}(z_{\tau})$ 
        \RETURN $z_{\tau-1}$
    \end{algorithmic}
\end{algorithm}

\begin{algorithm}[h!]
    \caption{SlidingWindow-Denoising}
    \label{alg:sliding_window_denoising_fusion}
    \textbf{Input}: Latent sequence $z_{\tau} \in \mathbb{R}^{c \times F \times h \times w}$ of length $F$, a keyframe set $\mathcal{K} = \{K_1, K_2, \ldots, K_n\}$, window size $w$, stride $s$ \\
    \textbf{Output}: Sliding window denoised latent ${z}_{\tau-1}$
    
    \begin{algorithmic}[1]
        \STATE $\mathcal{Z} \gets \emptyset$
        \FOR {$i = 1$ to $n$}
            \STATE $\text{start} \gets (i-1) \cdot s$ 
            \STATE $z_{\tau}^{sub} \gets z_{\tau}[\text{start}:\text{start} + w - 1]$
            \STATE $z_{\tau-1}^{sub} \gets \text{Denoise}(z_{\tau}^{sub}, K_i)$
            \STATE $\mathcal{Z} \gets \mathcal{Z} \cup \{z_{\tau-1}^{sub}\}$ 
        \ENDFOR
        \STATE $z_{\tau-1} \gets \text{Fuse}(\mathcal{Z})$
    \end{algorithmic}
\end{algorithm}

\begin{algorithm}[ht!]
    \caption{DiffuseSlide}
    \label{alg:sliding_window_denoising_fusion}
    \textbf{Input}: Latent sequence $z_{\tau} \in \mathbb{R}^{c \times F \times h \times w}$ at timestep $\tau$, noise-reinjection step $\delta$ \\
    \textbf{Output}: Denoised latent ${z}_0$
    
    \begin{algorithmic}[1]
        \WHILE{denoising timestep $\tau > 0$}
        \IF{$\tau > \delta$}
           \STATE $z_{\tau-1} = \text{SlidingWindow-Denoise-Reinject}(z_{\tau})$
        \ELSE
           \STATE $z_{\tau-1} = \text{SlidingWindow-Denoise}(z_{\tau})$ 
        \ENDIF
        \STATE ${\tau} = {\tau-1} $
        \ENDWHILE
        \RETURN ${z}_0$
    \end{algorithmic}
\end{algorithm}

\begin{figure}[ht!]
    \centering
    \includegraphics[width=1.0\columnwidth]{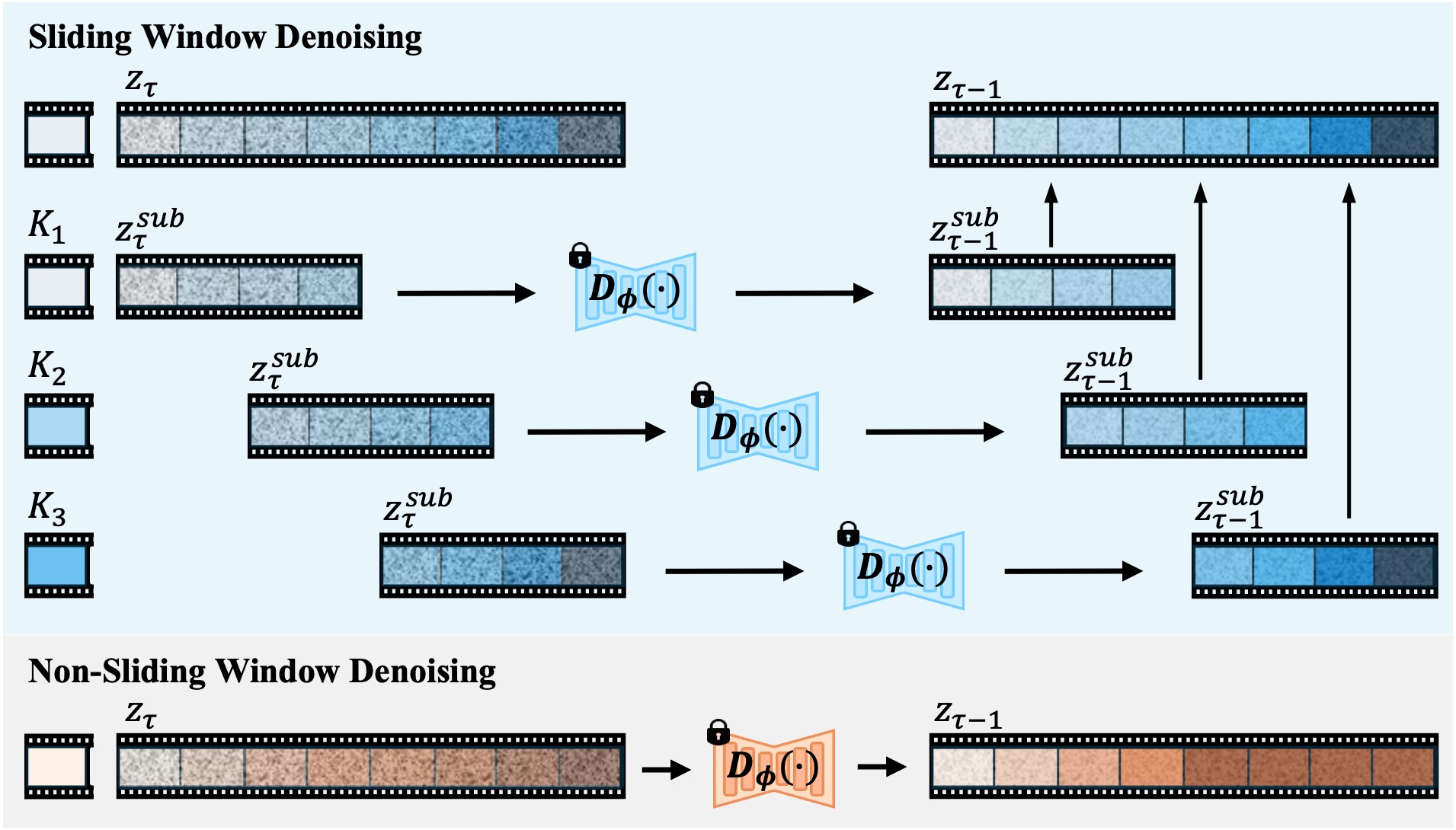} 
    \caption{Direct denoising of the full sequence latent may result in insufficient conditioning for the latter frames, as conditioning is limited to the initial frame. To address this, we introduce sliding window denoising technique that segments the video latent into subsequences, each conditioned on its respective keyframe. After denoising each subsequence, overlapping frames are averaged to reconstruct the full latent sequence at each timestep.}
    \vspace{-5.0mm}
    \label{fig:sliding_window}
\end{figure}

\subsection{Sliding Window for Multi-Image Conditioning}
\vspace{-1mm}
Our proposed noise-denoise and noise re-injection strategy effectively transforms linearly interpolated frames into a temporally aligned video while preserving the overall flow of keyframes. However, the extended latent with interpolated frames often exceeds the capacity of the pre-trained image-to-video model. This limitation stems from the inherent constraints of the video model, which was originally optimized for handling shorter sequences. Most image-to-video models apply positional encoding relative to the first frame, and this encoding combines with the diffusion time embedding as input to the U-Net. When the sequence length of frames surpasses the capacity learned by the pre-trained model, the resulting frames begin to exhibit blurriness, structural inconsistencies, and loss of fine details.

To address this, previous approaches, including ZeroSmooth~\cite{yang2024zerosmooth}, have adopted a noise rescheduling method and windowed attention mechanism that is initially introduced in FreeNoise~\cite{qiu2023freenoise}. However, a significant limitation remains: due to the architecture of image-to-video models, where each denoising step can only condition on a single image, these methods cannot utilize multiple keyframes as conditions throughout the process. This results in inefficiencies, as only the first frame is used for cross-attention conditioning, which inherently limits its influence on the latter frames in a long sequence. As the sequence progresses, the connection to the first frame naturally weakens, reducing the effectiveness of conditioning from the first frame in generating later frames that align with the desired keyframe structure.

To overcome this limitation, we propose segmenting the extended sequence into multiple subsequences using a sliding window approach. By structuring each subsequence starting from a keyframe, we allow each subsequence to condition directly on its corresponding keyframe during denoising. This design enables us to denoise each subsequence independently, aligning the interpolated frames more closely with the keyframes and resulting in a more temporally consistent and high-quality latent across the entire video.

By employing these methods, we can fully utilize the capacity of the base model, effectively reducing flickering and ghosting artifacts and generating high frame-rate videos without blurring or over-saturation. The overall pipeline of \textit{DiffuseSlide} is illustrated in \cref{fig:fig_pipeline}.
\vspace{-3mm}
\begin{figure*}[ht!]
    \centering
    \includegraphics[width=1.0\textwidth]{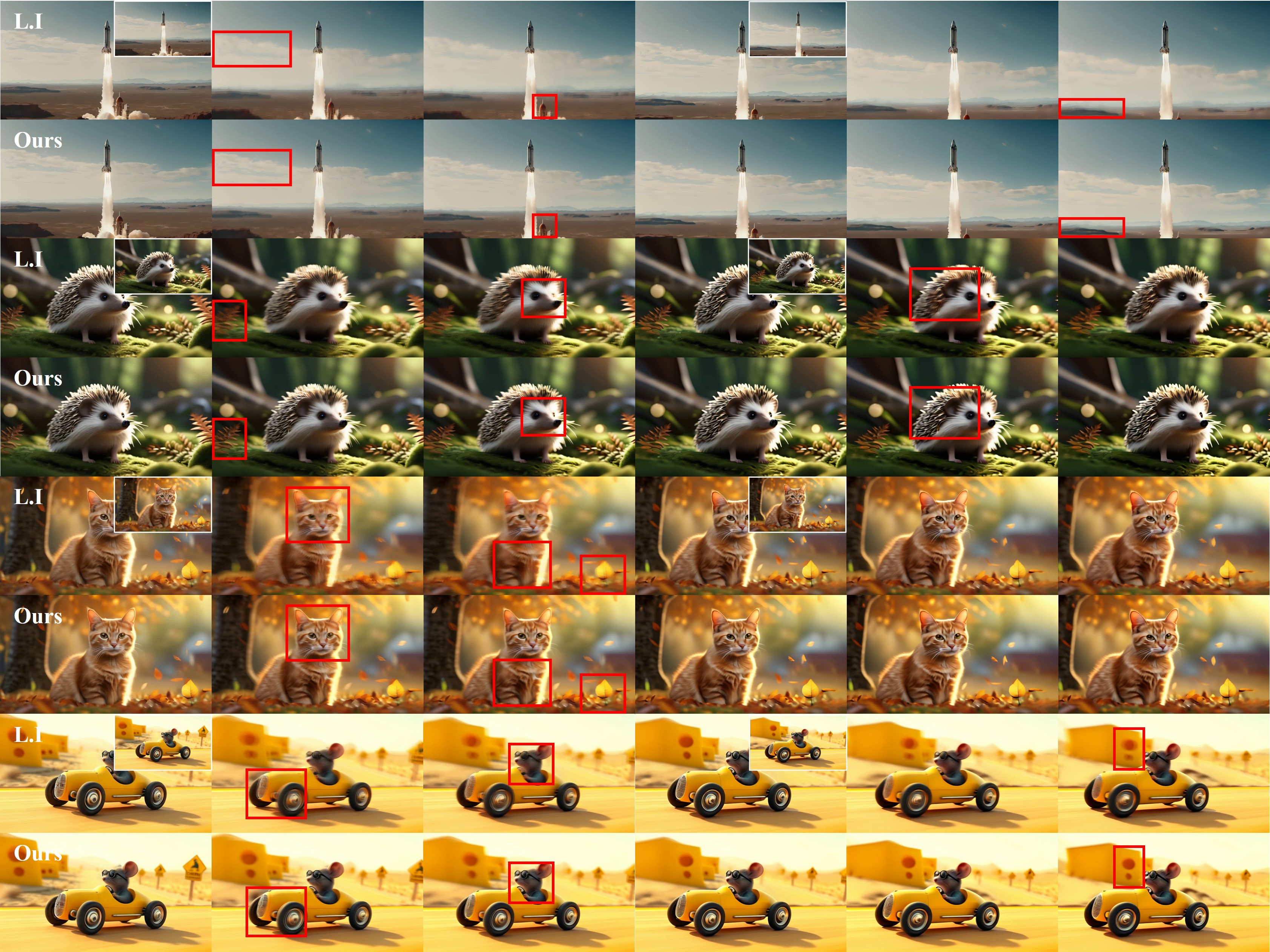} 
    \vspace*{-5mm}
    \caption{\textbf{Qualitative results showcasing high frame-rate video generation.} The images illustrate the smooth transitions and artifact reduction achieved with DiffuseSlide in 4x frame-rate experiments. ``L.I'' refers to Linear Interpolation.}
    \label{fig:main_qualitative}
\end{figure*}

\vspace{-1mm}
\section{Experiments}
\vspace{-1mm}

\subsection{Experiments Setup}
\vspace{-1mm}
In our experiments, we utilized two different image-to-video models: Stable Video Diffusion (SVD) and I2VGen-XL. Both models generate key frames from a single input image as a condition before applying our proposed DiffuseSlide pipeline.
For key frame generation, SVD produces 14 key frames with a resolution of 576 × 1024, and I2VGen-XL generates 16 key frames at a resolution of 704 × 1280.
Once the key frames were generated, we applied DiffuseSlide to both models to interpolate and generate high frame-rate videos with 2× and 4× frame-rate expansions. This approach does not require any additional training or fine-tuning of the base models, making it an efficient and adaptable solution for high frame-rate video generation.

\subsection{Evaluation}
\vspace{-1mm}
For our image-to-video generation experiments, we use the WebVid-10M~\cite{bain2021frozen} dataset, which consists of over 10 million high-definition videos across a wide range of scenarios. We randomly sampled 2,048 video clips from the dataset, as the specific video instances mentioned in~\cite{yang2024zerosmooth} are not publicly available, and used the first frame of each video as the reference image for generation. The generated videos maintain the same resolution as the original key frames and are expanded using our \textit{DiffuseSlide} pipeline to achieve 2× and 4× frame-rate increases.

We evaluate the quality of the generated videos using Fréchet Video Distance (FVD)~\cite{unterthiner2018towards}. 
Moreover, we measure the Peak Signal-to-Noise Ratio (PSNR) and the Structural Similarity Index (SSIM) between the initially generated low frame-rate video (key frames) and the corresponding frames in the high frame-rate video, to evaluate their similarity. Note that PSNR and SSIM do not directly assess the quality of the generated high frame-rate video, but rather the changes introduced during the high frame-rate video generation.
% Moreover, to evaluate the similarity between the initially generated low frame-rate video (key frames) and the corresponding frames in the high frame-rate video, we employ the Peak Signal-to-Noise Ratio (PSNR) and the Structural Similarity Index (SSIM).

\begin{table*}[ht!]
    \vspace*{-2mm}
    \scriptsize 
    \centering
    \resizebox{0.8\textwidth}{!}{
        \begin{tabular}{ c | c | c c c | c c c}
            \toprule
                \multirow{2}{*}{Model} & \multirow{2}{*}{Method} & \multicolumn{3}{c|}{$2\times$} & \multicolumn{3}{c}{$4\times$} \\ 
                
              &  & FVD($\downarrow$) & PSNR($\uparrow$) & SSIM($\uparrow$)  & FVD($\downarrow$) & PSNR($\uparrow$) & SSIM($\uparrow$) \\
            \midrule
            \multirow{4}{*}{SVD} 
                & $\text{Direct Inference}$ & 1079.26 & 19.18 & 0.638 & 1175.38 & 16.79 & 0.627  \\
                & $\text{ZeroSmooth}^\dagger$ & 779.6 & 30.92 & 0.927 & 752.1 & 30.74 & 0.921  \\
                & $\text{Linear Interp.}$ & 874.17 & - & - & 900.26  & - & -  \\
                & $\text{DiffuseSlide}$ & \textbf{584.63} & \textbf{32.52} & \textbf{0.932} & \textbf{636.41} & \textbf{32.38} & \textbf{0.931}\\
            \midrule
            \multirow{3}{*}{I2VGen-XL}
                & $\text{Direct Inference}$ & 1175.13 & 16.20 & 0.562 & 1396.52 & 14.74 & 0.486 \\
                & $\text{Linear Interp.}$ & 1098.61 & - & - & 1164.83  & - & -  \\
                & $\text{DiffuseSlide}$ & \textbf{965.58} & \textbf{39.83} & \textbf{0.984} & \textbf{998.82} & \textbf{40.10} & \textbf{0.984}\\  
            \bottomrule
        \end{tabular}
    }
    \vspace{-1mm}
    \caption{\textbf{Quantitative results of high frame-rate generation experiments.} Direct inference refers to generating all frames directly without interpolation or additional conditioning. For the ZeroSmooth baseline, we used the values reported in their paper, as the official code was unavailable. The best results are highlighted in \textbf{bold} for clarity.}
    \label{tab:tab_main}
    \vspace*{-5mm}
\end{table*}

\begin{figure*}[ht!]
    \centering
    \includegraphics[width=1.0\textwidth]{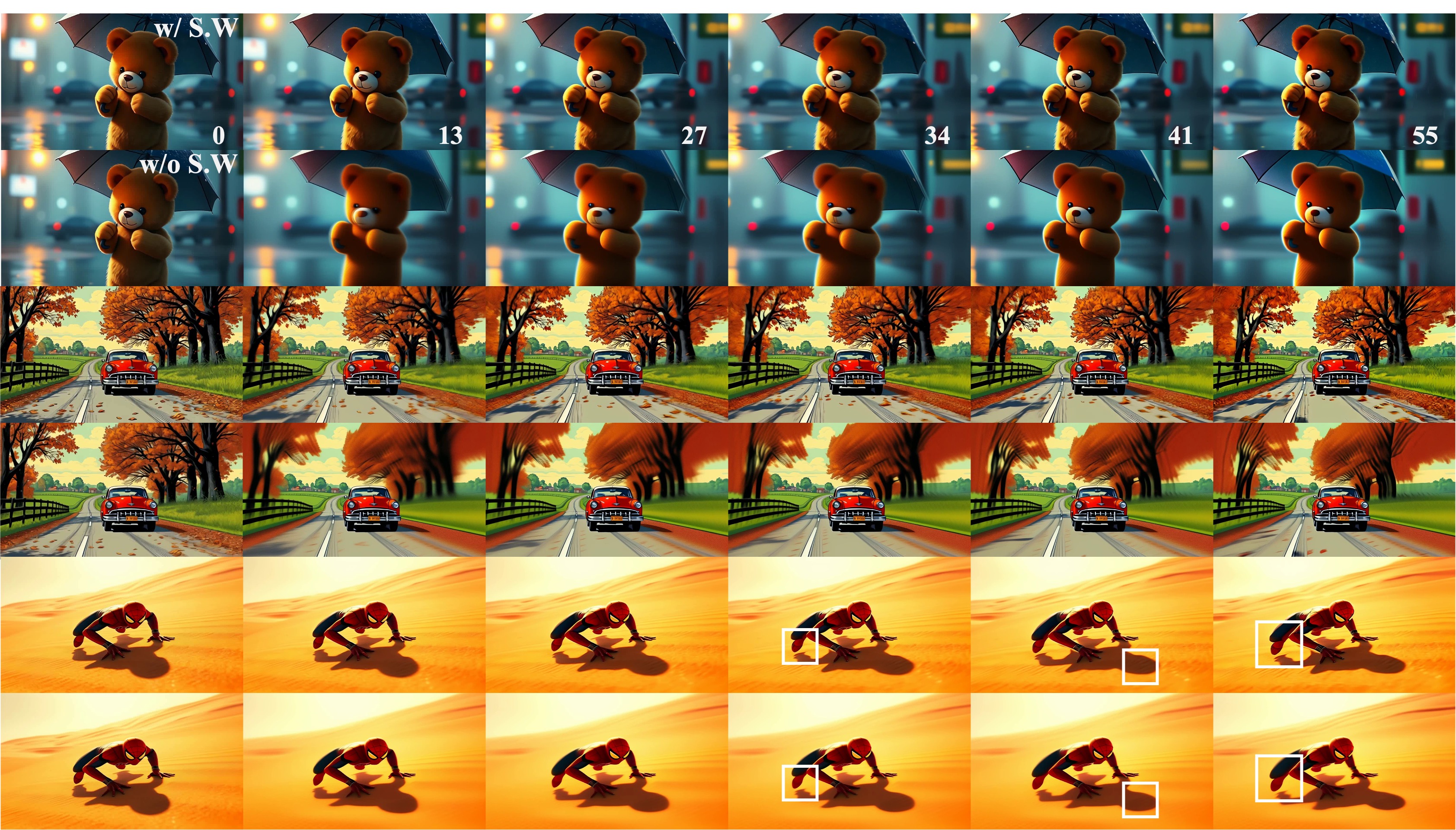} 
    \vspace{-5mm}
    \caption{\textbf{Ablation study results showing the impact of sliding window denoising.} The quality improvement across frames demonstrates the effectiveness of these components in preserving temporal consistency and reducing degradation in later frames. ``S.W" refers to Sliding Window denoising and the number on the bottom left in each frame denotes the frame number.}
    \label{fig:sw_ablation}
\end{figure*}

\begin{table}[ht!]
    \scriptsize
    \centering
    \resizebox{0.45\textwidth}{!}{
        \begin{tabular}{c|c|c c c}
            \toprule
                \multicolumn{2}{c|}{Method} & \multicolumn{3}{c}{$2\times$} \\
            \midrule
                \text{S.W} & \text{N.R.I} & FVD($\downarrow$) & PSNR($\uparrow$) & SSIM($\uparrow$) \\
            \midrule
                \textcolor[rgb]{0.91,0.3,0.24}{\ding{55}} & \textcolor[rgb]{0.91,0.3,0.24}{\ding{55}} & 1079.26 & 19.18 & 0.638 \\
                \textcolor[rgb]{0.91,0.3,0.24}{\ding{55}} & \textcolor[rgb]{0.18,0.8,0.45}{\ding{51}} & 618.03 & 28.80 & 0.882 \\
                \textcolor[rgb]{0.18,0.8,0.45}{\ding{51}} & \textcolor[rgb]{0.91,0.3,0.24}{\ding{55}} & 866.39 & \textbf{32.52} & 0.929 \\
                \textcolor[rgb]{0.18,0.8,0.45}{\ding{51}} & \textcolor[rgb]{0.18,0.8,0.45}{\ding{51}} & \textbf{584.63} & \textbf{32.52} & \textbf{0.932} \\
            \midrule
                \multicolumn{2}{c|}{Method} & \multicolumn{3}{c}{$4\times$} \\
            \midrule
                \text{S.W} & \text{N.R.I} & FVD($\downarrow$) & PSNR($\uparrow$) & SSIM($\uparrow$) \\
            \midrule
                \textcolor[rgb]{0.91,0.3,0.24}{\ding{55}} & \textcolor[rgb]{0.91,0.3,0.24}{\ding{55}} & 1175.38 & 16.79 & 0.627 \\
                \textcolor[rgb]{0.91,0.3,0.24}{\ding{55}} & \textcolor[rgb]{0.18,0.8,0.45}{\ding{51}} & 786.52 & 26.98 & 0.852 \\
                \textcolor[rgb]{0.18,0.8,0.45}{\ding{51}} & \textcolor[rgb]{0.91,0.3,0.24}{\ding{55}} & 735.02 & \textbf{32.72} & \textbf{0.931} \\
                \textcolor[rgb]{0.18,0.8,0.45}{\ding{51}} & \textcolor[rgb]{0.18,0.8,0.45}{\ding{51}} & \textbf{636.41} & 32.38 & \textbf{0.931} \\
            \bottomrule
        \end{tabular}
    }
    
    \caption{\textbf{Quantitative results of the ablation study on DiffuseSlide.} 
    We analyze the impact of two key components: Sliding Window Denoising (S.W) and Noise Re-Injection (N.R.I). The symbols \textcolor[rgb]{0.18,0.8,0.45}{\ding{51}} and \textcolor[rgb]{0.91,0.3,0.24}{\ding{55}} indicate that the corresponding component is enabled or disabled, respectively. The best results are highlighted in \textbf{bold} for clarity.}
    \vspace{-5mm}
    \label{tab:tab_ablation}
\end{table}

\begin{figure}[ht!]
    \centering
    \includegraphics[width=0.95\columnwidth]{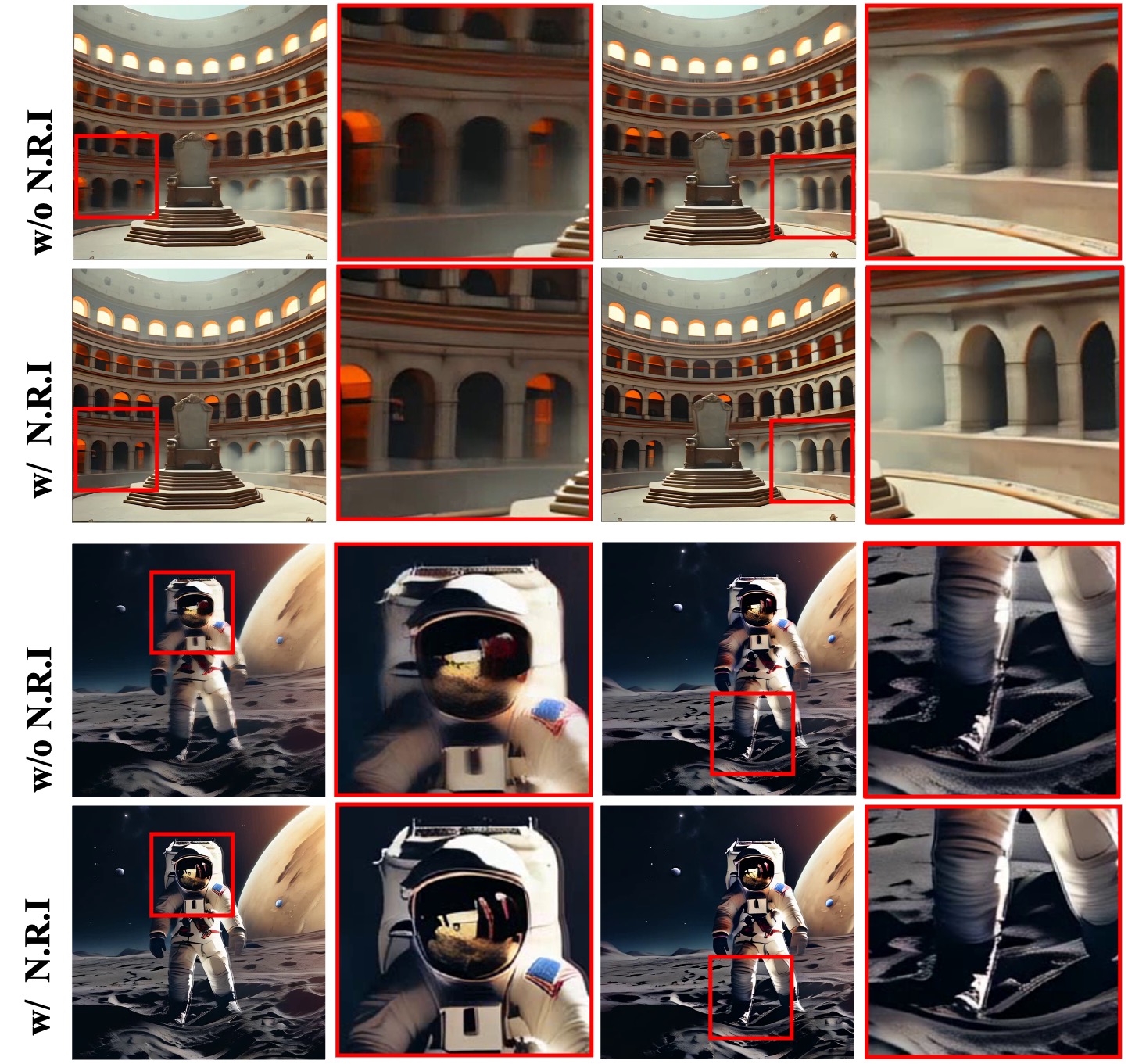} 
  
    \caption{Ablation study results showing the effect of Noise Re-Injection (``N.R.I" denotes Noise Re-Injection).
    This ablation study highlights the artifact reduction and temporal consistency improvements achieved through our Noise Re-Injection technique. The results demonstrate that Noise Re-Injection effectively realigns interpolated frames with the original data manifold, thereby enhancing both frame-to-frame coherence and visual quality in fast-motion sequences. ``N.R.I" denotes Noise Re-Injection.
    }
    \label{fig:nri_ablation}

\end{figure}
\vspace{-1mm}

\subsection{Result}
\vspace{-1mm}
\paragraph{Qualitative results.}
For qualitative results, we compare our method, \textit{DiffuseSlide}, with Linear Interpolation (L.I). For Linear Interpolation, 14 frames are generated using the SVD model~\cite{blattmann2023stable} and then expanded by interpolating additional frames. In \textit{DiffuseSlide}, frames are generated following the pipeline introduced in the main paper. Although we considered including ZeroSmooth~\cite{yang2024zerosmooth} as a training-free baseline method, its code is unavailable, which prevents us from comparing qualitative results. The qualitative comparisons are presented in \cref{fig:main_qualitative}.

\vspace{-4.5mm}
\paragraph{Quantitative results.}
We present the quantitative evaluation results in \cref{tab:tab_main}. As shown in the table, our proposed pipeline achieves the highest performance across all key metrics, including FVD, PSNR, and SSIM. These results demonstrate the effectiveness of our method in generating high frame-rate videos with superior temporal and spatial consistency.

In particular, the lower FVD score highlights the improved realism and overall video quality achieved by our approach, while the higher PSNR and SSIM values indicate that our pipeline maintains greater fidelity to the original key frames compared to the other baselines, preserving both details and structures. This is especially important in high frame-rate video generation, where maintaining the integrity of key frames while generating smooth transitions between them is crucial.

These quantitative results confirm that our method not only enables the generation of high frame-rate videos with minimal flickering or artifacts but also ensures that key frames are accurately referenced. Compared to baseline methods, which suffer from repetitive motion artifacts and flickering issues, our approach consistently delivers smoother, more visually coherent video outputs.

\subsection{Ablation Studies}
\vspace{-1.0mm}
We conducted ablation studies to validate the effectiveness of two key components: Noise Re-Injection (N.R.I) and Sliding Window denoising (S.W), both of which play a crucial role in enhancing temporal consistency and overall video quality. The results of these studies are summarized in Table~\ref{tab:tab_ablation}.
\vspace{-2.0mm}

\paragraph{Noise Re-Injection.}  
Noise Re-Injection iteratively introduces noise after each denoising step, effectively mitigating artifacts from interpolated frames. Without noise re-injection, the model struggles to reach the smooth data manifold, resulting in lower quality (Table~\ref{tab:tab_ablation}). Noise re-injection progressively realigns frames toward the data manifold, significantly improving temporal consistency.  

The visual impact of Noise re-injection is further illustrated in Fig.~\ref{fig:nri_ablation}. Without N.R.I, interpolated frames exhibit noticeable artifacts and inconsistencies, particularly in high-motion regions (highlighted in red boxes). By reintroducing noise at each step, our method effectively refines frame transitions, reducing flickering and preserving fine details. This results in smoother motion and improved perceptual quality, demonstrating the necessity of noise re-injection in high frame-rate video generation.

\vspace{-6mm}
\paragraph{Sliding Window denoising.}
We further evaluated the necessity of sliding window denoising. Without it, video latents conditioned solely on the initial frame degrade as the sequence progresses, causing blurriness and loss of fidelity (Fig.~\ref{fig:sw_ablation}). Sliding window denoising resolves this by independently conditioning subsequences on corresponding keyframes, maintaining consistent quality across frames.

Combining both techniques (S.W and N.R.I) achieves optimal results, confirming their complementary roles in high frame-rate video generation by ensuring both structural consistency and smooth temporal transitions, while effectively reducing artifacts and flickering.

\begin{table}[ht!]
    \centering
    \scriptsize 
    \resizebox{0.45 \textwidth}{!}{
        \begin{tabular}{c|c c c c}
            \toprule
                \multirow{2}{*}{Method} & \multicolumn{4}{c}{$2\times$} \\  
                 & FVD($\downarrow$) & PSNR($\uparrow$) & SSIM($\uparrow$) & Time \\
            \midrule
                $\text{LDMVFI}$ & 684.98 & \textbf{39.42} & \textbf{0.977} & 73 sec \\
                $\text{DiffuseSlide}$ & \textbf{584.63} & 32.52 & 0.932 & 273 sec \\
            \midrule
                \multirow{2}{*}{Method} & \multicolumn{4}{c}{$4\times$} \\ 
                 & FVD($\downarrow$) & PSNR($\uparrow$) & SSIM($\uparrow$) & Time \\
            \midrule
                $\text{LDMVFI}$ & 701.62 & \textbf{34.91} & \textbf{0.960} & 224 sec \\
                $\text{DiffuseSlide}$ & \textbf{636.41} & 32.38 & 0.931 & 268 sec \\
            \bottomrule
        \end{tabular}
        \vspace{-7mm}
    }
    \caption{\textbf{Quantitative comparison between DiffuseSlide and the Training-based method}. The best results are highlighted in \textbf{bold} for clarity.}
    \label{tab:tab_trainig_based}

\end{table}

\vspace{-5mm}
\subsection{Comparison to Training-based Methods}
\vspace{-1.0mm}
We further compare our approach, \textit{DiffuseSlide}, against the LDMVFI~\cite{danier2024ldmvfi}, which is a training-based method. Table~\ref{tab:tab_trainig_based} presents quantitative results for frame-rate expansions.

Although LDMVFI achieves higher PSNR and SSIM scores, these metrics primarily prioritize exact reconstruction of key frames rather than overall perceptual video quality. DiffuseSlide achieves superior performance in terms of Frechet Video Distance (FVD), highlighting better realism and temporal consistency. Despite its higher inference cost at the 2× setting, DiffuseSlide becomes computationally competitive at 4×. 

Furthermore, the training-free nature of \textit{DiffuseSlide} provides a significant advantage in real-world applications where training or fine-tuning large-scale video diffusion models is impractical. This makes our method a compelling choice for scenarios that require high frame-rate video generation without the computational overhead of training-based approaches.
\vspace{-3mm}
\section{Limitation and Discussion}
\vspace{-1mm}
Current open-source image-to-video models are optimized for short sequences. Extending them to longer videos without additional training requires either continuous attention operations or multiple inferences using techniques like our sliding window approach. However, the inherently slow inference of diffusion models further extends generation times. Advancements in diffusion sampling speeds or models specifically trained for longer sequences could further improve efficiency.

Additionally, as \textit{DiffuseSlide} is a training-free approach, its performance is inherently dependent on the capabilities of large-scale pre-trained video diffusion models. Current open-source video diffusion models still have limitations in handling extremely large motion or highly dynamic scenes, which may lead to minor inconsistencies in complex scenarios. However, as video diffusion models continue to improve, these limitations are expected to be naturally mitigated, further enhancing the effectiveness of our method.
\vspace{-2mm}
\section{Conclusion}
\vspace{-1mm}
In this work, we presented \textit{DiffuseSlide}, a novel, training-free approach for high frame-rate video generation using pre-trained image-to-video diffusion models. Our method introduces a pipeline that leverages low frame-rate videos as keyframes and performs interpolation through noise re-injection and sliding window latent denoising, enabling the generation of smooth, high-quality videos without additional model training or fine-tuning. By preserving both temporal and spatial consistency, \textit{DiffuseSlide} effectively mitigates challenges such as flickering and ghosting artifacts commonly observed in linear interpolation.

Through extensive experiments, we demonstrated the effectiveness of \textit{DiffuseSlide}, achieving competitive performance across key video quality metrics, including FVD, PSNR, and SSIM. Our approach sets a new benchmark on the WebVid-10M dataset, surpassing existing baselines and proving its suitability for high frame-rate video generation in resource-constrained environments, such as gaming, VR/AR, and video streaming.

\textit{DiffuseSlide} provides a valuable tool to generate high-quality, high frame-rate videos, paving the way for smoother user experiences across various applications. Future research may focus on optimizing diffusion models for faster sampling or developing models specifically trained for high frame-rate sequences to further enhance both performance and efficiency.

{
    \small
    \bibliographystyle{ieeenat_fullname}
    \bibliography{main}
}

\clearpage
\appendix
\section{Linear Interpolation}
The interpolation is performed in latent space. The frame between two key frames is computed as:
\begin{equation}
    z_t = \alpha z_{t-1} + (1 - \alpha) z_{t+1}
\end{equation}

\noindent where $\alpha$ depends on the relative time position. In the $2\times$ setting, one frame is inserted between each consecutive frames, while three frames are inserted in the $4\times$ setting. For the final frame, we duplicate it once in $2\times$ interpolation and three times in $4\times$.

\section{Implementation Details}
As the image-to-video backbone, we utilized both the Stable Video Diffusion (SVD) and the I2VGen-XL models. SVD is optimized to generate 14-frame videos, while I2VGen-XL produces 16-frame outputs. 

For the diffusion sampling process, we used the Euler scheduler with 25 steps for SVD and 50 steps for I2VGen-XL. The hyperparameters were consistently set as follows: noising step $\tau = 8$ (SVD) and $\tau = 15$ (I2VGen-XL), noise re-injection step $\delta = 3$ (SVD) and $\delta = 4$ (I2VGen-XL), and the number of iterations $M = 5$ for both models. 

For model-specific configurations, frames per second (fps) and decode chunk size were fixed at 7 and 1 for SVD and I2VGen-XL, respectively. The motion bucket ID in SVD was initially set at 128 for low frame-rate video generation and subsequently reduced to 64 and 32 for 2× and 4× frame-rate expansion tasks. 
All experiments were performed on an NVIDIA RTX 4090 GPU.

\section{Evaluation Metrics}
To evaluate the performance of our method, we employ three key metrics: Frechet Video Distance (FVD), Peak Signal-to-Noise Ratio (PSNR), and Structural Similarity Index (SSIM). 

Since our task focuses on high frame-rate video generation, FVD is the most critical metric, as it effectively assesses the smoothness, temporal consistency, and overall visual quality of the generated videos. To compute FVD, we sample 2,048 videos from the WebVid-10M dataset and compare the generated high frame-rate videos against their original counterparts.

In addition to FVD, we use PSNR and SSIM to measure how well keyframes are preserved when extending low frame-rate videos to 2× and 4× frame rates. These metrics quantify the fidelity of generated frames relative to the original keyframes. However, in the case of Linear Interpolation, keyframes remain unchanged when expanding videos to 2× or 4× frame rates, making it impossible to compute PSNR and SSIM meaningfully in such cases.

\section{Effect Number of Iterations}
In Table \ref{tab:tab_nri_2}, we investigate the effect of varying the number of Noise Re-Injection iterations ($M$), a key hyperparameter in our pipeline, in the $4\times$ frame-rate scenario. Although increasing $M$ from 5 to 7 slightly reduces FVD, we observe that the PSNR and SSIM metrics favor $M=5$, suggesting better preservation of keyframe details. Moreover, selecting $M=5$ strikes a balance between maintaining temporal consistency and computational efficiency, making it the optimal choice for our pipeline.

\begin{table}[ht!]
    \centering
    \scriptsize
    \resizebox{0.45 \textwidth}{!}{
        \begin{tabular}{c|c c c c}
            \toprule
            & \multicolumn{4}{c}{$4\times$} \\
            \hline
               $\text{M}$ & $\text{FVD}(\downarrow)$ & $\text{PSNR}(\uparrow)$ & $\text{SSIM}(\uparrow)$ & $\text{time(sec)}(\downarrow)$\\
            \midrule
                $5$ & 636.41 & \textbf{32.38} & \textbf{0.931} & \textbf{162.13} \\
                $7$ & \textbf{626.62} & 32.14 & 0.929 & 194.26\\
            \bottomrule
        \end{tabular}
    }
    \caption{\textbf{Quantitative results of the ablation study on the number of iterations $M$ for Noise Re-Injection in $4\times$ frame-rate scenarios.} The best results are highlighted in \textbf{bold} for clarity. Inference time was measured by averaging the time required to generate 20 videos using a single NVIDIA A100 GPU.}
    \label{tab:tab_nri_2}
\end{table}

\begin{figure}[ht!]
    \centering
    \includegraphics[width=0.95\columnwidth]{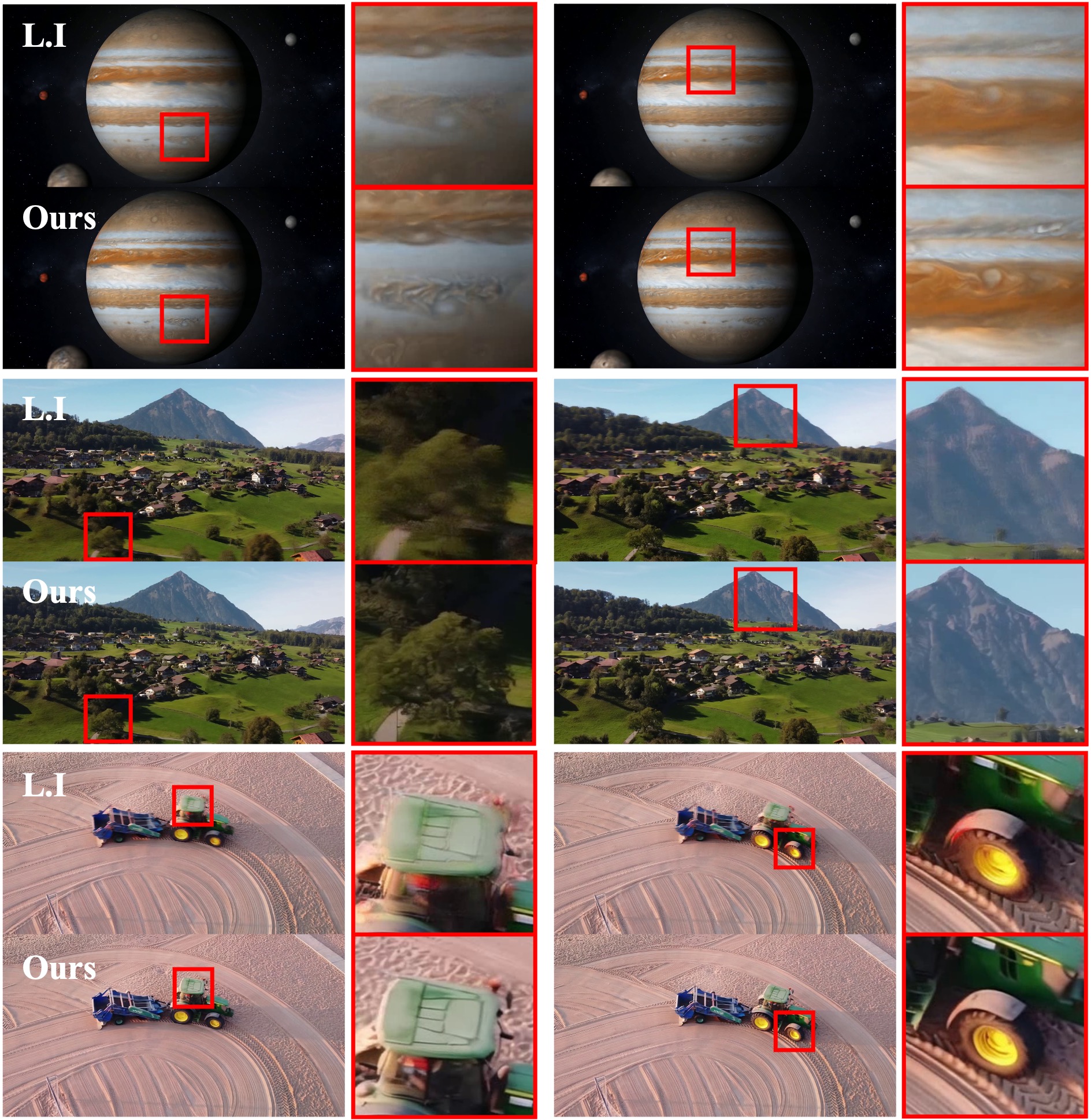} 
    \caption{\textbf{Detailed qualitative results for video-to-video generation.} This figure highlights the detailed improvements in smoothness and artifact reduction achieved by DiffuseSlide in $4\times$ frame-rate experiments.}
    \label{fig:supple_video_to_video_qualitative_zoom}
\end{figure}

\section{Video-to-Video Generation}
\textit{DiffuseSlide} is a versatile framework that excels in both image-to-video and video-to-video generation. It seamlessly transforms single images into coherent video sequences and enhances low frame-rate input videos into high frame-rate outputs, delivering high-quality results for a wide range of tasks.

\begin{figure*}[t]
    \centering
    \includegraphics[width=1.0\textwidth]{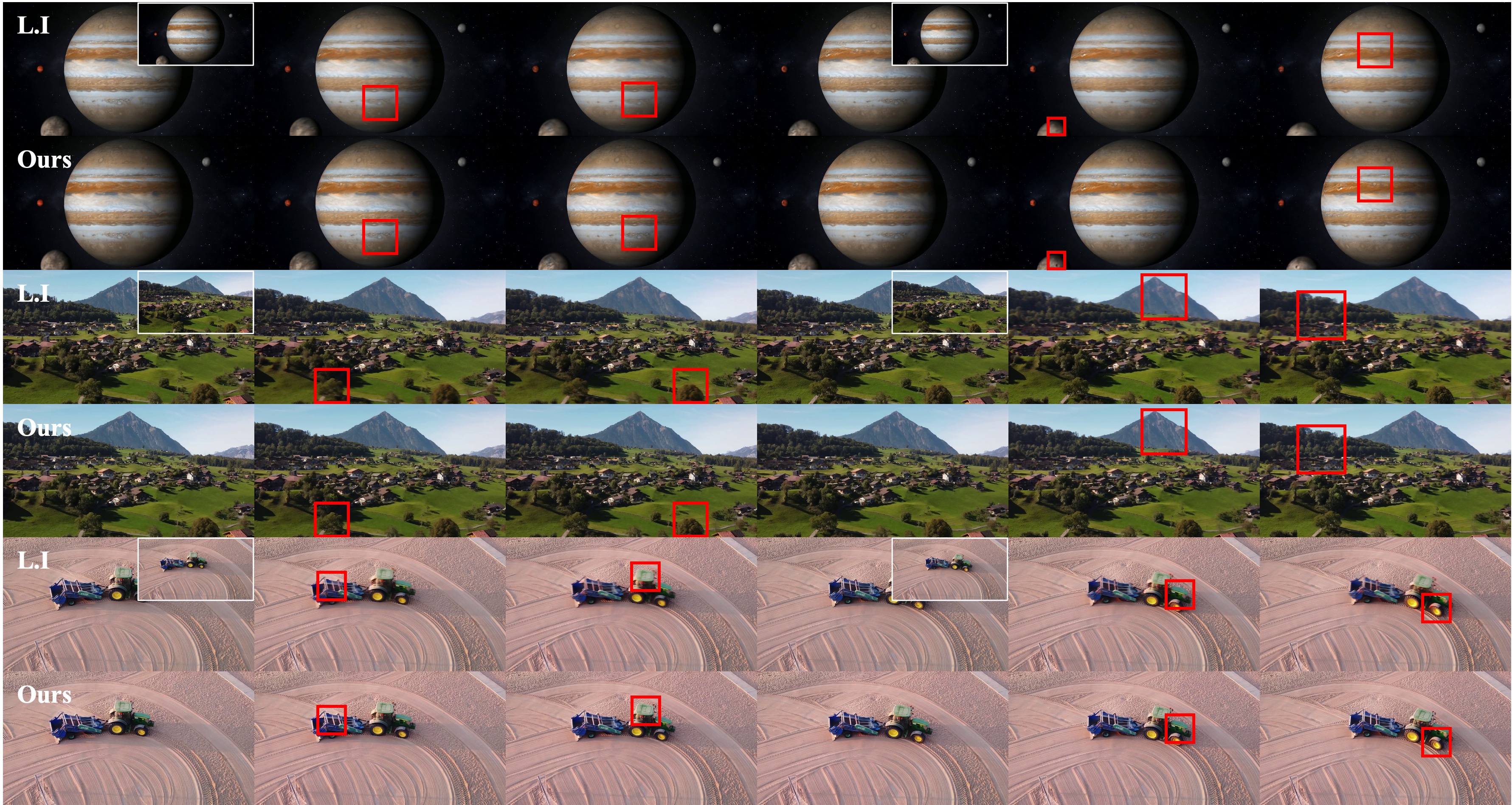} 
    \caption{\textbf{Overall qualitative results for video-to-video generation.} The figure illustrates the enhanced temporal consistency and artifact reduction achieved with DiffuseSlide in $4\times$ frame-rate video-to-video scenarios. ``L.I'' refers to Linear Interpolation. See Figure \ref{fig:supple_video_to_video_qualitative_zoom} for detailed comparisons.}
    \label{fig:supple_video_to_video_qualitative}
    \vspace{-20mm}
\end{figure*}

 By leveraging the Sliding Window Denoising technique, our method efficiently processes long input videos without being constrained by GPU memory limitations. This approach ensures temporally consistent and visually high-quality outputs, even for extended sequences. We used the Pexels dataset\footnote{\url{https://www.pexels.com}} and sampled frames at intervals of 4 to use them as input videos. Figures \ref{fig:supple_video_to_video_qualitative_zoom}, \ref{fig:supple_video_to_video_qualitative} illustrate the superior performance of DiffuseSlide compared to baseline approaches, such as Linear Interpolation (L.I).

\section{Additional Qualitative Result}
\begin{figure}[ht!]
    \centering
    \includegraphics[width=1.0\columnwidth]{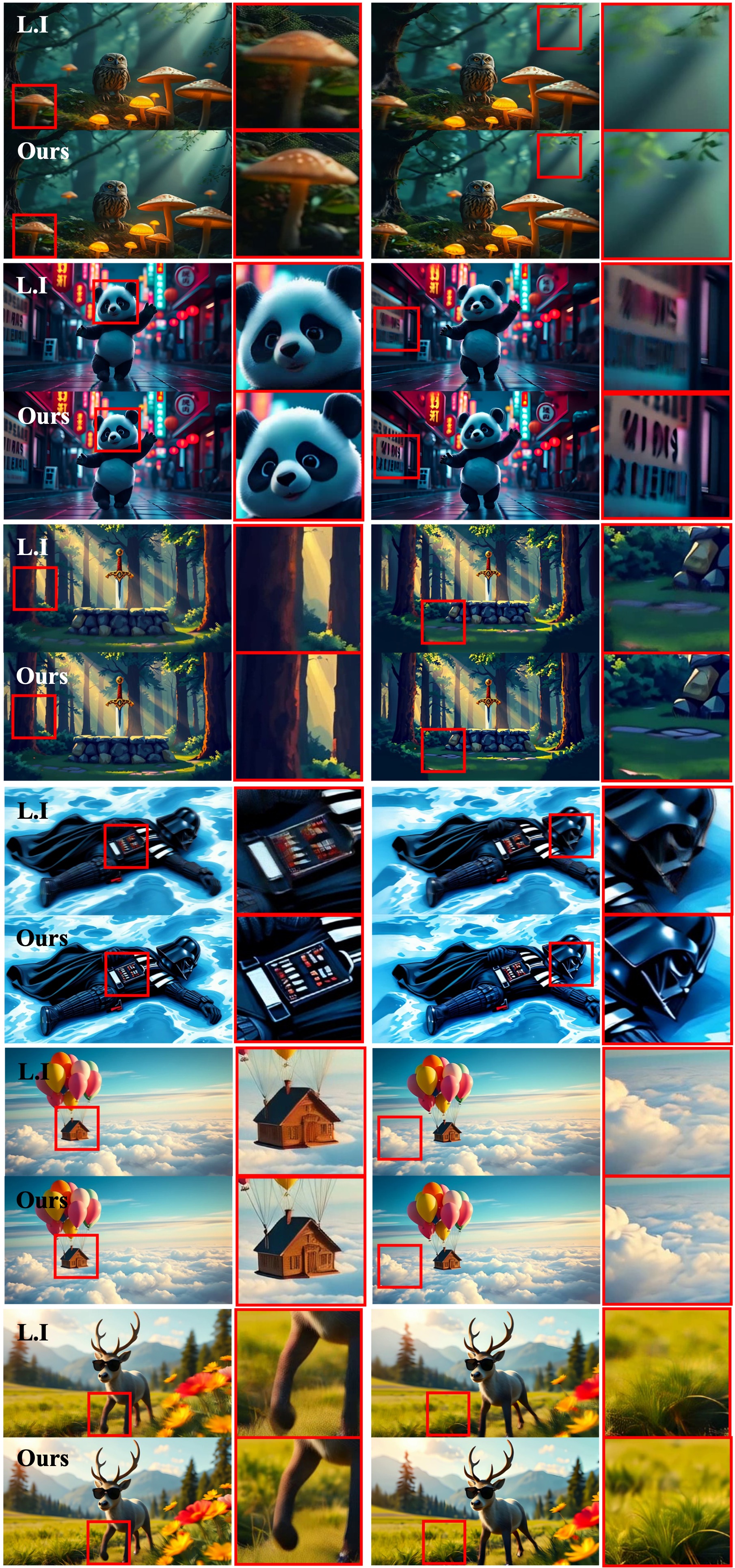} 
    \caption{\textbf{Detailed improvements in high frame-rate video generation.} The figure showcases the effectiveness of DiffuseSlide in reducing artifacts and enhancing smooth transitions for $4\times$ frame-rate generation.}
    \label{fig:supple_main_qualitative_zoom}
\end{figure}

\begin{figure*}[t]
    \centering
    \includegraphics[width=1.0\textwidth]{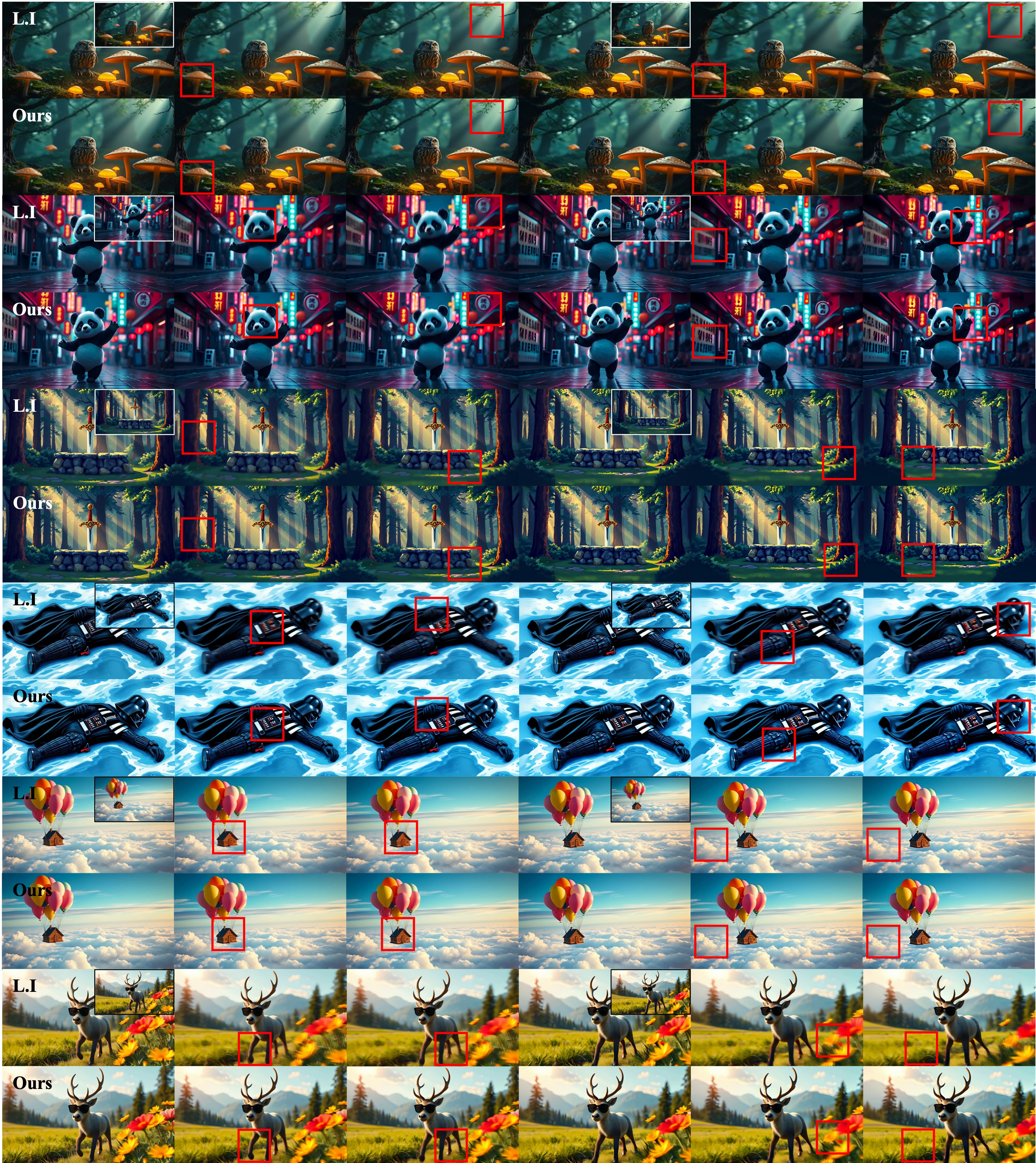} 
    \caption{\textbf{Overall qualitative results for high frame-rate video generation.} This figure demonstrates the smooth transitions and reduced artifacts achieved with DiffuseSlide in $4\times$ frame-rate scenarios. ``L.I'' refers to Linear Interpolation. For a closer look at specific details, see Figure \ref{fig:supple_main_qualitative_zoom}.}
    \label{fig:supple_main_qualitative}
    \vspace{-20mm}
\end{figure*}

This section presents the qualitative improvements achieved by DiffuseSlide in high frame-rate video generation. Compared to the Linear Interpolation (L.I) method, DiffuseSlide demonstrates superior temporal consistency and reduces artifacts significantly. Figures \ref{fig:supple_main_qualitative_zoom}, \ref{fig:supple_main_qualitative} illustrate these results.

\vspace{-10pt}

\vspace{5mm}

\end{document}